\documentclass[a4paper]{article}

\usepackage{INTERSPEECH2021}
\usepackage{color}
\title{Improving Mandarin Speech Recogntion with Block-augmented Transformer}
\name{Xiaoming Ren$^1$,Huifeng Zhu$^1$, Liuwei Wei$^1$, Minghui Wu$^1$, Jie Hao$^1$}
\address{
  $^1$Mininglamp Technology, Beijing, China}
\email{\{renxiaoming,zhuhuifeng,weiliuwei,wuminghui,haojie\}@mininglamp.com}

\begin{document}

\maketitle

\begin{abstract}
  Recently Convolution-augmented Transformer (Conformer)\cite{gulati2020conformer} has shown promising results in Automatic Speech Recognition (ASR), outperforming the previous best published Transformer Transducer\cite{Zhang2020TransformerTA}. In this work, we believe that the output information of each block in the encoder and decoder is not completely inclusive, in other words, their output information may be complementary. We study how to take advantage of the complementary information of each block in a parameter-efficient way, and it is expected that this may lead to more robust performance. Therefore we propose the Block-augmented Transformer for speech recognition, named Blockformer. We have implemented two block ensemble methods: the base Weighted Sum of the Blocks Output (Base-WSBO), and the Squeeze-and-Excitation module\cite{Hu2020SqueezeandExcitationN} to Weighted Sum of the Blocks Output (SE-WSBO). Experiments have proved that the Blockformer significantly outperforms the state-of-the-art Conformer-based models on AISHELL-1, our model achieves a CER of 4.29\% without using a language model and 4.05\% with an external language model on the testset. 
\end{abstract}
\noindent\textbf{Index Terms}: speech recognition, weight, blockformer, conformer block, transformer block, Squeeze-and-Excitation module

\section{Introduction}
Deep Learning has been applied successfully to Automatic
Speech Recognition (ASR) \cite{Hinton2012DeepNN}. A variety of neural network architectures for acoustic modeling have been explored. For example, DNNs \cite{Dahl2012ContextDependentPD}, CNNs \cite{Sainath2013DeepCN}, RNNs \cite{Graves2013SpeechRW} and end-to-end models\cite{Graves2014TowardsES,Chan2016ListenAA,Bahdanau2016EndtoendAL,Chiu2018StateoftheArtSR}. 
Currently, there are mainly three E2E models: Neural Transducer (NT) \cite{gulati2020conformer,Graves2013SpeechRW,Graves2012SequenceTW,Rao2017ExploringAD} models, Attention-Based Encoder-Decoder (AED) models \cite{Chan2016ListenAA,Bahdanau2016EndtoendAL,Dong2018SpeechTransformerAN} and Connectionist Temporal Classification \cite{Graves2006ConnectionistTC} (CTC) models \cite{Graves2014TowardsES,Kriman2020QuartznetDA}. These E2E models treat ASR as a sequence-to-sequence task that directly learns speech to text mapping with a neural network. 
The NT model consists of an encoder, which maps input acoustic frames into a higher-level representation, and a prediction and joint network which together correspond to the decoder network\cite{Rao2017ExploringAD}. The decoder is conditioned on the history of previous predictions. The NT training is unstable and takes more memory which may limit the training speed.
The AED model is composed of an encoder, which encodes acoustic features, and a decoder, which generates a sentence. The architectures of many state-of-the-art ASR systems \cite{Karita2019ImprovingTE} are based on the AED models. However, the AED model outputs token by token, where each token depends on previously generated tokens and acoustic context, causing recognition delays. 
On the other hand, the CTC model contains only an encoder and outputs all tokens independently. Although its decoding speed is faster than the AED model, in term of recognition accuracy it is generally inferior due to the conditional independence assumption between output tokens. However, in real environment speech recognition tasks, the attention model performs poorly because the alignment estimated in the attention mechanism is easily corrupted by the noise. 

In recent years, there are some researches for joint CTC-attention model \cite{Kim2017JointCB,Zhang2020UnifiedSA} which use a shared-encoder representation trained by both CTC and attention model objectives simultaneously within the multi-task learning framework. In this paper, we focus on the joint CTC-attention models, aiming at better performance by using more model output information.

\begin{figure}[t]
  \centering
  \includegraphics[width=\linewidth]{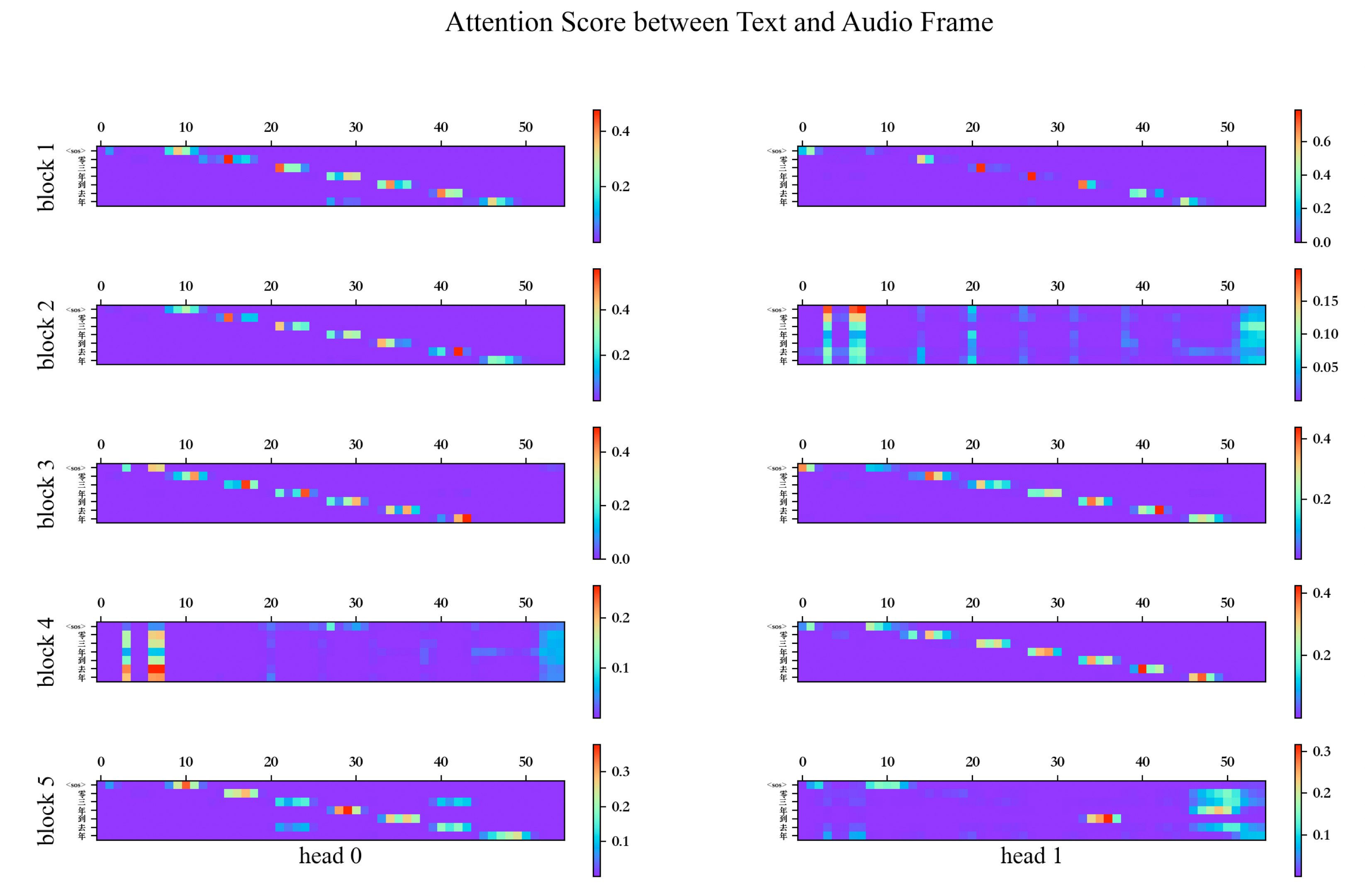}
  \caption{Attention information visualization between different layers and heads, indicates the diversity of information, but also complementarity.}
  \label{fig:speech_production}
\end{figure}

In our work\footnote{https://github.com/Mininglamp-Technology/ASR-BlockFormer}, we analyze the contribution of different layers and heads through attention visualization, as depicted in Fig 1. We consider that the diversity of information between blocks is useful. We study how to take advantage of the output information of each block. We believe that the output information of each block in the encoder and decoder is not completely inclusive, and perhaps complementary. Combine the output of each block efficiently to make full use of complementary information. 
Motivated by this, we introduce a novel model, named Blockformer, achieves state-of-the-art results on Aishell-1. The Blockformer adds a block ensemble module, which exploits the output information of each block in a parameter-efficient way.  
We have implemented two block ensemble methods: the base Weighted Sum of the Blocks Output (Base-WSBO), and the Squeeze-and-Excitation module\cite{Hu2020SqueezeandExcitationN} to Weighted Sum of the Blocks Output (SE-WSBO).
Our experiments are mainly conducted on a public Mandarin Chinese dataset AISHELL-1. Results show that a CER of 4.29\% without using a language model and 4.05\% with an external language model on the testset.

The rest of the paper is organized as follows. Section 2 presents
various components of the Blockformer model in detail. Section 3 and Section 4 present our experimental settings, detailed results and ablation study. Concluding remarks and a discussion of future directions are presented in Section 5.

\section{Blockformer}
In this section, we introduce the Blockformer. As shown in Figure 2, the proposed model is built upon the typical Attention-Based Encoder-Decoders (AEDs). The Blockformer encoder uses the Conformer block, and the decoder uses the Transformer block. The conformer block we used is the same as \cite{gulati2020conformer}. In the transformer block, we use the relative positional encoding instead of the absolute positional encoding. Block ensemble is proposed in this paper, which is used in encoder and decoder respectively. Moreover, we implement two Block ensemble methods. It will be described in detail below.

\subsection{Encoder}
Our audio encoder first processes speech features with a convolution
subsampling layer followed by several conformer blocks. The encoder is composed of a stack of identical Conformer blocks.
\subsubsection{Conformer block}

We use the standard conformer block\cite{gulati2020conformer}, which is composed of four modules stacked together, i.e, a feed-forward module, a self-attention module, a convolution module, and a second feed-forward module, the two Feed Forward modules sandwiching the Multi-Headed Self-Attention module and the Convolution module. Mathematically, this means, for input \(\mathbf{x}_{i}\)
to a Conformer block \(i\), the output \(\mathbf{y}_{i}\) of the block is:
\begin{equation}
  \widetilde{\mathbf{x}}_{i} = LN(\mathbf{x}_{i} + \frac{1}{2}FFN(\mathbf{x}_{i}))
  \label{eq1}
\end{equation}
\begin{equation}
  {\mathbf{x}_{i}}\\' = LN(\widetilde{\mathbf{x}}_{i} + MHSA(\widetilde{\mathbf{x}}_{i}))
  \label{eq2}
\end{equation}
\begin{equation}
  {\mathbf{x}_{i}}\\'\\' = LN({\mathbf{x}_{i}}\\' + Conv({\mathbf{x}_{i}}\\'))
  \label{eq3}
\end{equation}
\begin{equation}
  {\mathbf{y}_{i}} = LN({\mathbf{x}_{i}}\\'\\' + \frac{1}{2}FFN({\mathbf{x}_{i}}\\'\\'))
  \label{eq4}
\end{equation}
where FFN refers to the Feed-Forward module, MHSA refers to the Multi-Head Self-Attention module, Conv refers to the
Convolution module, and LN refers to the Layernorm module .

\subsubsection{Relative positional encoding}
We employ the relative sinusoidal positional encoding scheme which is an important technique from Transformer-XL\cite{Dai2019TransformerXLAL}. The relative positional encoding allows the self-attention module to generalize better on different input lengths so that the resulting encoder is more robust to the variance of the utterance length.

\begin{figure}[t]
  \centering
  \includegraphics[width=\linewidth]{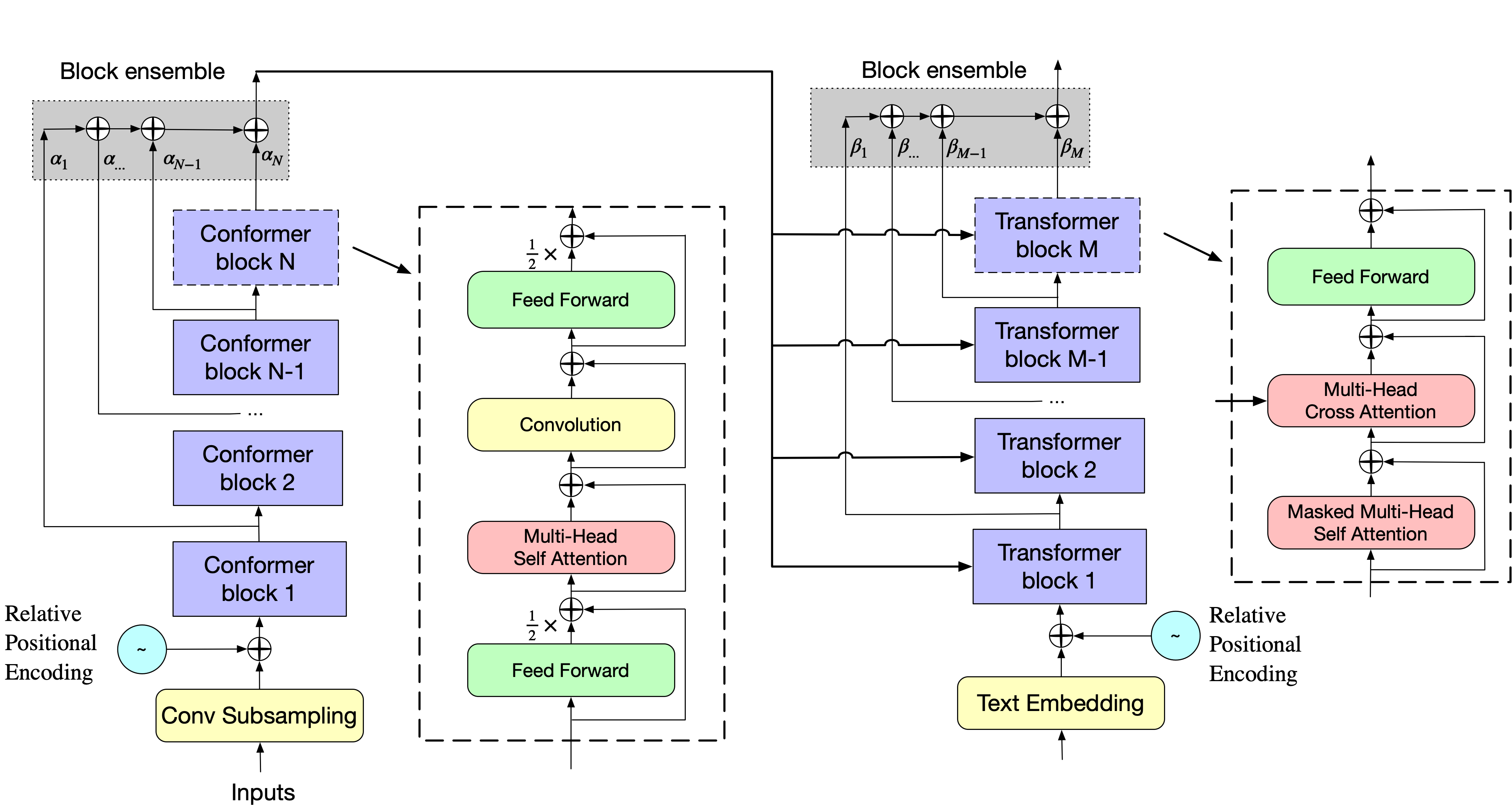}
  \caption{Blockformer model architecture with Base-WSBO block ensemble .}
  \label{fig:speech_production}
\end{figure}

\begin{figure}[t]
  \centering
  \includegraphics[width=\linewidth]{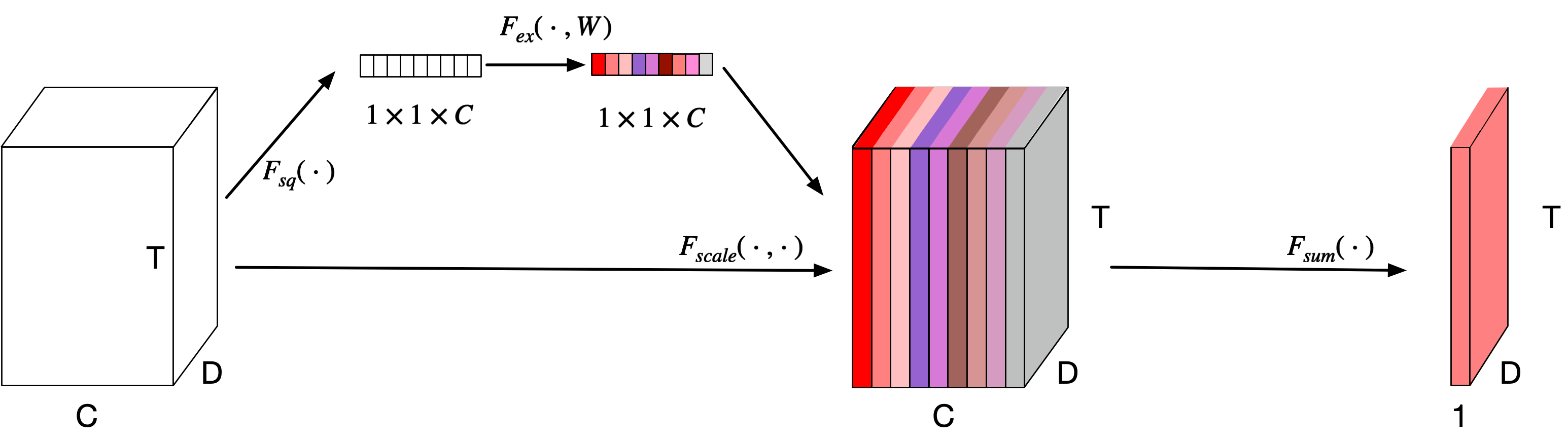}
  \caption{SE-WSBO}
  \label{fig:speech_production}
\end{figure}

\subsubsection{Block ensemble}
We have implemented two block ensemble methods. The first method is very intuitive which is the base Weighted Sum of the Blocks Output (Base-WSBO). The structure is depicted in Figure 2. The weight of the block output is a learnable parameter. The second method is the Squeeze-and-Excitation module \cite{Hu2020SqueezeandExcitationN} to Weighted Sum of the Blocks Output (SE-WSBO). The structure is depicted in Figure~\ref{fig:speech_production}. We can improve the quality of representations produced by the SE module\cite{Hu2020SqueezeandExcitationN} by explicitly modeling the interdependencies between the output of all blocks. We allow the network to perform feature recalibration, through which it can learn to use global information to selectively emphasize informative block output and suppress unuseful ones.

\noindent\textbf{Base-WSBO.} \ The base Weighted Sum of the Blocks Output, \( \mathbf{y}_{i} \) is the block output, \( \alpha_{i} \) is a scalar .
\begin{equation}
  \widetilde{\mathbf{y}} = \sum_{i=1}^{N}\hat{\alpha}_{i}\mathbf{y}_{i}
  \label{eq3}
\end{equation}
\begin{equation}
  \hat{\alpha}_{i} = \frac{e^{\alpha_{i}}}{\sum_{j}e^{\alpha_{j}}}  = Softmax(\alpha_{i})
  \label{eq3}
\end{equation}
where in order to make the sum of the coefficients equal to 1, we optionally use softmax to constrain the coefficient \( \alpha_{i} \).

\noindent\textbf{SE-WSBO.}\ Since Base-WSBO does not fully establish the correlation between blocks, each output \(\mathbf{y}_{i}\) of the
block is unable to exploit contextual information outside of this region. In order to tackle the issue of exploiting blocks dependencies, we first consider how to extract each block output feature. We squeeze global block output information into a channel descriptor. This is achieved by using global average pooling to generate channel-wise statistics. The c-th element of \( \mathbf{z}\) is calculated by:
\begin{equation}
  z_{c} = \mathbf{F}_{sq}(\mathbf{y}_{c}) = \frac{1}{T\times D}\sum_{i=1}^{T}\sum_{j=1}^{D}\mathbf{y}_{c}(i,j)
  \label{eq3}
\end{equation}
where \(\mathbf{y}_{c}\in \mathbb{R}^{T\times D}\) refers to the c-th block output, T and D are the dimensions.
\begin{equation}
  \mathbf{s} = \mathbf{F}_{ex}(\mathbf{z},\mathbf{W}) = \sigma(\mathbf{W}_{2}\delta (\mathbf{W}_{1}\mathbf{z}))
  \label{eq3}
\end{equation}
where \(\sigma\) refers to the sigmoid function, \(\delta\) refers to the ReLU \cite{Nair2010RectifiedLU} function,
\( \mathbf{W}_{1}\in \mathbb{R}^{\frac{C}{r}\times C}\)
and
\( \mathbf{W}_{2}\in \mathbb{R}^{C\times \frac{C}{r}}\). \( \frac{C}{r}\) is the bottleneck dim. Usually, we set \(C_{encoder}=12, C_{decoder}=6, r=1\).
\begin{equation}
  \widetilde{\mathbf{y}}_{c} = \mathbf{F}_{scale}(\mathbf{y}_{c},s_{c}) = s_{c}\mathbf{y}_{c}
  \label{eq3}
\end{equation}
where \(\mathbf{F}_{scale}\) refers to
channel-wise multiplication by the scalar \(s_{c}\).
\begin{equation}
  \widetilde{\mathbf{y}} = \mathbf{F}_{sum}(\widetilde{\mathbf{y}}_{c}) = \sum_{c=1}^{N}\widetilde{\mathbf{y}}_{c}
  \label{eq3}
\end{equation}
where \(\widetilde{\mathbf{y}}\) is the final block ensemble output .

\subsection{ Attention based decoder}
The decoder is also composed of a stack of identical Transformer blocks. In addition to the two modules (a feed-forward module, a self-attention module) , the decoder inserts a third module, named multi-Head Cross-Attention module (MHCA) which performs multi-head attention over the output of encoder block ensemble. Similar to the encoder, we employ Relative positional encoding and Block ensemble.
\subsubsection{Transformer block}
For input \(\mathbf{x}_{i}\)
to the decoder Transformer block i, the output \(\mathbf{y}_{i}\) of the block is:
\begin{equation}
  {\mathbf{x}_{i}}\\' = LN(\mathbf{x}_{i} + MHSA(\mathbf{x}_{i})
  \label{eq1})
\end{equation}
\begin{equation}
  {\mathbf{x}_{i}}\\'\\' = LN({\mathbf{x}_{i}}\\' + MHCA({\mathbf{x}_{i}}\\',\widetilde{\mathbf{y}}))
  \label{eq3}
\end{equation}
\begin{equation}
  {\mathbf{y}_{i}} = LN({\mathbf{x}_{i}}\\'\\' + FFN({\mathbf{x}_{i}}\\'\\'))
  \label{eq4}
\end{equation}
where \(\widetilde{\mathbf{y}} \) refers to the encoder output after block ensemble.

\subsection{ Hybrid CTC-Attention Objective}
With the aim to take advantage of ctc and attention, the CTC
and attention loss can be combined \cite{Watanabe2017HybridCA}. Both CTC and attention-based methods have their own drawbacks. CTC often has poor results due to the assumption of conditional independence between output tokens. Since the estimated alignment in the attention mechanism is easily corrupted by noise, the performance of the attention model is also generally poor.

The combination of these two not only helps the model to converge, but also enables the model to take full advantage of token dependencies.
The hybrid CTC-Attention objective is defined in Equation 14, where x is the acoustic feature, y is the corresponding annotation. \(L_{CTC}(x, y)\), \( L_{AED}(x, y) \)are the CTC and AED loss respectively, \(\lambda\in (0,1)\) is a hyperparameter which balance the importance of CTC and AED loss:

\begin{equation}
  L_{hybrid} = \lambda L_{CTC}(\mathbf{x},\mathbf{y}) + (1-\lambda)L_{AED}(\mathbf{x},\mathbf{y})
  \label{eq5}
\end{equation}

\section{Experimental setting}

\subsection{Datasets}
In this paper, we validated the proposed two Blockformer methods (Base-WSBO and SE-WSBO) on two Mandarin speech recognition datasets: public AISHELL-1 corpus \cite{Bu2017AISHELL1AO}, internal 1400 hours corpus.
The AISHELL-1 corpus consists of 178 hours of labeled speech collected from 400 speaker with high fidelity microphone. We construct the decoding graph TLG using its annotated text as a corpus for language model. The 1400-hour internal corpus is collected from some service domains, such as medicine, cosmetic and estate, which is more diverse in data and more challenging in speech recognition. For the AISHELL-1 task, we use the 150-hour for training and the 18-hour development set for early-stopping. 
The character error rate (CER\%) is reported in the 7176-sentence test set (about 10 hours). 
For the 1400-hour internal corpus, we use a 1400-hour for training, a 6-hour development set for early-stopping, and a 13-hour test set for evaluation.

\subsection{Experimental Setup}
For all experiments, the input features are 80-dimensional
log Mel-filterbank(FBank) computed on 25ms window with
10ms shift. We use the open-source WeNet toolkit\cite{Yao2021WeNetPO} to build both the vanilla hybrid CTC/attention Conformer baseline and our proposed Blockformer. We use SpecAugment\cite{2019,Park2020SpecaugmentOL} for data augmentation with the frequency mask parameter (F = 10), the time mask parameter(T=50), and the number of frequency and time masks(mF = mT = 2). We choose 4233 and 4599 characters (including \(\langle pad\rangle, \langle eos\rangle, \langle sos\rangle \) labels) as model units for
AISHELL-1, 1400-hour internal corpus respectively.

We build the baseline model with a 12-layer encoder and
a 6-layer decoder following the WeNet recipe\cite{Yao2021WeNetPO}.
We employ h = 4 parallel attention heads in the blockformer models. For every layer, we use \(d_{k}=d_{v}=d_{model}/h=64\), \(d_{ffn}= 2048\) . The base model has about 46M parameters. Our Base-WSBO model only adds some scaler weight( Increased number of parameters: 18 ), SE-WSBO model adds SE-module( Increased number of parameters: 360 ).

AdamOptimizer\cite{Vaswani2017AttentionIA} is used 
with \(learning\_rate = 0.002, warm\_up = 50000\), and gradient
clipping at 5.0. Moreover, we employ label smoothing of value \(\epsilon_{ls} = 0.1 \)\cite{Szegedy2016RethinkingTI} and dropout rate of \(P_{drop} = 0.1\).
For regularization, we apply dropout \cite{Srivastava2014DropoutAS} in each residual unit of the conformer block and transformer block, i.e, to the output of each module, before it is added to the module input. We set the weight \( \lambda \) of the CTC branch during joint training to 0.3. During joint decoding, we set the CTC-weight \( \lambda \) to 0.5. We also train an external n-gram LM followed by WeNet recipe\cite{Yao2021WeNetPO}. To avoid overfitting, we averaged the 30 best model parameters in the development dataset. For the optimal combination of SE-WSBO acoustic model and language model TLG for Aishell-1 testset, the parameters detail as follows: \(acoustic\_scale = 3.26,ngram\_weight=0.93,lattice\_beam=16,beam=32\).

Furthermore, we use the gradient accumulation\cite{Hermans2017AccumulatedGN,gradient-accumulation} during training, where the gradients are updated every 4 batches.
The baseline dataloader sorts all the utterances according to the frames length, packs them in sequence by the batch size, and randomly select to be passed to the model. The benefit of sorting all utterances makes training more efficient, which can occupy as much GPU memory as possible. The disadvantage is that the package of each batch is fixed, which may not allow the model to learn a better combination of information. In view of the above shortcomings, our batch data are selected from all utterances instead of packages. We train models using 4 Nvidia A100 GPUs for at most 120 epochs with a batch size of 20. 

\section{Experimental results}
We firstly present our results on the Aishell-1 test dataset to provide a deep insight into our method. The effectiveness of the proposed method is further verified on the larger corpus (1400-hours internal corpus).
To evaluate the effectiveness of the two Blockformer methods Base-WSBO and SE-WSBO, we conduct some experiments to compare differences. The performance of the models is evaluated based on character error rates (CERs) both without and with external language models. All our experimental results are based on the attention-rescore two-step decoding method \cite{Sainath2019TwoPassES, Zhang2020UnifiedSA}.

\subsection{Results of Aishell-1}
\begin{table}[t]
  \caption{Experimental results on the Aishell-1 test dataset (CER\%) }
  \label{tab:word_styles2}
  \centering
  \begin{tabular}{lll}
    \toprule
    \textbf{Method}  & \textbf{No LM}& \textbf{With LM}  \\
    \midrule
    \textbf{AEDs}(\emph{previous work}) & \\
    Espnet\cite{Watanabe2018ESPnetES}    &  4.90\%   & 4.70\%              \\
    WeNet\cite{Yao2021WeNetPO}     & 4.61\% & 4.36\%                 \\
    K2\cite{k2}        & ~~  -- &4.26\%            \\  
    \midrule
    \textbf{NTs}(\emph{previous work}) & \\
    Neural Transducer+LFMMI\cite{Tian2021ConsistentTA}  & ~~ -- &4.18\%           \\ 
    \midrule
    \textbf{Blockformer}(\emph{our work}) & \\
    Base-WSBO  & 4.48\% & 4.22\%                  \\
    Base-WSBO+Softmax  & 4.54\% & 4.28\%                  \\
    SE-WSBO     & 4.29\% & \textbf{4.05}\%                  \\
    \bottomrule
  \end{tabular}
\end{table}

Table 1 compares the (CER) result of our model on the Aishell-1 test dataset with a few public models include: Espnet\cite{Watanabe2018ESPnetES}, WeNet\cite{Yao2021WeNetPO}, K2\cite{k2} and
Neural Transducer+LFMMI\cite{Tian2021ConsistentTA}. The first three models are all AED model structures, and the last is NT based. All our evaluation results round up to 2 digit after decimal point. 
From Table 1, we can see that our proposed Base-WSBO model is better than previous three AEDs model, but it is still a little worse compared with the NT model. In the experiment, we additionally learn that the effect of adding softmax to Base WSBO became worse. This case may indicate that the fixed sum of coefficients is not effective.

Without a language model, the performance of our
SE-WSBO model already achieves competitive results of CER 4.29\%
on testset outperforming the known WeNet model.  With
the language model, our model achieves the lowest CER 4.05\% among all the existing models. This clearly demonstrates the effectiveness of block ensemble in a single neural network.

\subsection{Ablation Study}
Table 2 shows the impact of each change to SE-WSBO on the Aishell-1 test dataset. We perform experiments to study the effect of varying number of blocks at block ensemble. E12D6 means utilizing all 12 block output information of the encoder and all 6 block output information of the decoder. E5D5 means use the last 5 block output of encoder and the ones of decoder respectively.   

We can see the major performance drops in Table 2 (1) replacing E12D6 with E5D5, which reduce the number of encoder block ensemble more than the one of the decoder. Comparison (4) can also clearly see the importance of the encoder block ensemble. The result of (2) shows that the use of relative position encoding in the decoder is not more effective than absolute position encoding. It may be that this replacement only affects the attention decoding result. In the two-step decoding method, the CTC is the first pass decoding, and the attention is only used for revision in the second pass. So the replacement of the positional encoding method on the decoder does not show the advantage. As can be seen from the results of experiment (3), using the random data reading method does not seem to be very critical, which may be affected by random factors. To verify the importance of encoder and decoder in block ensemble respectively, we design (4) and (5) experiments. The results show that the block ensemble of the encoder is more important than the one of decoder in SE-WSBO model. The experiments in (6) want to discuss how much performance can be improved by considering only relative position encoding in the decoder in combination with unsorted and unpacked training data.

\begin{table}[t]
  \caption{\noindent\textbf{Ablation study of the SE-WSBO (CER\%).} Starting from the SE-WSBO, we remove its features: (1)replacing E12D6 with E5D5; (2)replacing decoder self-attention with relative positional encoding with a vanilla self-attention layer \cite{Vaswani2017AttentionIA} with absolute positional encoding; (3)replacing the data preprocessing method used in SE-WSBO to randomly extract the training data one by one with which is packaged after the overall sorting of the training utterance length in the baseline; (4)replacing E12D6 with only E12; (5)replacing E12D6 with only D6; (6)without using E12D6. All ablation study results are evaluated on the Aishell-1 test dataset without the external LM.}
  \label{tab:word_styles2}
  \centering
  \begin{tabular}{lll}
    \toprule
    \textbf{Method} & \textbf{No LM} \\
    \midrule

    SE-WSBO     & 4.29\%                  \\
    \quad -- E12D6 + E5D5     & 4.48\%                  \\
    \quad\quad-- decoder Relative Pos + Abs Pos     & 4.49\% \\
    \quad\quad\quad-- no\_pack no\_sort + packsort    & 4.47\% \\
    \quad -- E12D6 + E12     & 4.39\%                  \\
    \quad -- E12D6 + D6     & 4.46\%                  \\
    \quad-- E12D6    & 4.52\% \\
    \bottomrule
  \end{tabular}
\end{table}

\subsection{Results of internal 1400 hours}
As in Table 3, our SE-WSBO model is still beneficial for internal 1400 hours training. The CER value of SE-WSBO model decreased by 3\% in the 13-hour testset.

Moreover, comparing the No LM and With LM columns in Table 2, we also observe that the performance of the language model improves slightly when the amount of training data increases. 
It is possible that the acoustic model has already learned enough knowledge of the language model in the large amount of training data.

\begin{table}[t]
  \caption{Experimental results on 1400-hour internal dataset (CER\%)  }
  \label{tab:word_styles2}
  \centering
  \begin{tabular}{lll}
    \toprule
    \textbf{Method} & \textbf{No LM} & \textbf{With LM} \\
    \midrule
    WeNet-baseline     & 23.49\%      & 23.06\% \\
    SE-WSBO           &  22.78\%      & 22.37\% \\
    \bottomrule
  \end{tabular}
\end{table}

\section{CONCLUSIONS AND FUTURE WORK}
In this work, we introduced the Blockformer, an architecture that
integrated block ensemble for end-to-end speech recognition. We studied and discussed two block ensemble methods, and demonstrated that it was beneficial to the performance of the Blockformer model.
The model achieved better results with a few extra parameters than
previous work on the Mandarin dataset Aishell-1, and achieved a new
state-of-the-art performance at 4.29\%/4.05\% for test dataset.
The Blockformer on a much larger and more challenging dataset also confirmed our findings. Our method was easy to implement and could also be applied to other models. 

There were some works we had not explored yet. For example, the decoder used in our experiment is still based on the transformer block. In fact, we did the experiment based on the conformer block, but the CER only dropped by about 3\%. We would like to write another paper to discuss this issue in detail in our future work.

\bibliographystyle{IEEEtran}
\bibliography{blockformer}

\end{document}